\documentclass[lettersize,journal]{IEEEtran}
\usepackage{amsmath,amsfonts}
\usepackage{algorithmic}
\usepackage{algorithm}
\usepackage{array}
\usepackage[caption=false,font=normalsize,labelfont=sf,textfont=sf]{subfig}
\usepackage{textcomp}
\usepackage{stfloats}
\usepackage{url}
\usepackage{verbatim}
\usepackage{graphicx}
\usepackage{cite}
\usepackage{booktabs}
\usepackage{multirow}
\usepackage{amssymb}   
\usepackage{makecell}  
\usepackage{siunitx} 
\usepackage[table]{xcolor}
\usepackage{orcidlink}
\hypersetup{hidelinks}

\begin{document}

\title{DiffPixelFormer: Differential Pixel-Aware Transformer for RGB-D Indoor Scene Segmentation}

\author{
Yan Gong\href{https://orcid.org/0000-0002-3148-8286}{\orcidlink{0000-0002-3148-8286}}, Jianli Lu\href{https://orcid.org/0000-0002-8459-9211}{\orcidlink{0000-0002-8459-9211}}, Yongsheng Gao*\href{https://orcid.org/0000-0002-1555-8328}{\orcidlink{0000-0002-1555-8328}}, Jie Zhao\href{https://orcid.org/0000-0002-6086-9387}{\orcidlink{0000-0002-6086-9387}},~\IEEEmembership{Senior Member,~IEEE}, Xiaojuan Zhang\href{https://orcid.org/0000-0002-7719-8890}{\orcidlink{0000-0002-7719-8890}},~\IEEEmembership{Senior Member,~IEEE}, and Susanto Rahardja\href{https://orcid.org/0000-0003-0831-6934}{\orcidlink{0000-0003-0831-6934}},~\IEEEmembership{Fellow,~IEEE}

\thanks{This work was supported by the National Outstanding Youth Science Fund Project of National Natural Science Foundation of China (Grant no. 52025054). \textit{(Corresponding author: Yongsheng Gao)}}

\thanks{Yan~Gong, Jianli~Lu, Yongsheng~Gao, and Jie~Zhao are with the State Key Laboratory of Robotics and System, Harbin Institute of Technology, Harbin 150001, China. (email: gongyan2020@foxmail.com, lujianli364@163.com, gaoys@hit.edu.cn, jzhao@hit.edu.cn).}
\thanks{Xiaojuan Zhang is with the Institute for Infocomm Research, A*STAR, Singapore. (email: xiaojuanzhang@ieee.org).}
\thanks{Susanto Rahardja is with the College of Information Science and Electronic Engineering, Zhejiang University, Hangzhou 310027, China (e-mail: susantorahardja@ieee.org).}
}

\markboth{IEEE Transactions on Multimedia}%
{IEEE Transactions on Multimedia}


\maketitle

\begin{abstract}
Indoor semantic segmentation is fundamental to computer vision and robotics, supporting applications such as autonomous navigation, augmented reality, and smart environments.
Although RGB-D fusion leverages complementary appearance and geometric cues, existing methods often depend on computationally intensive cross-attention mechanisms and insufficiently model intra- and inter-modal feature relationships, resulting in imprecise feature alignment and limited discriminative representation.
To address these challenges, we propose DiffPixelFormer, a differential pixel-aware Transformer for RGB-D indoor scene segmentation that simultaneously enhances intra-modal representations and models inter-modal interactions.
At its core, the Intra-Inter Modal Interaction Block (IIMIB) captures intra-modal long-range dependencies via self-attention and models inter-modal interactions with the Differential–Shared Inter-Modal (DSIM) module to disentangle modality-specific and shared cues, enabling fine-grained, pixel-level cross-modal alignment. Furthermore, a dynamic fusion strategy balances modality contributions and fully exploits RGB-D information according to scene characteristics.
Extensive experiments on the SUN RGB-D and NYUDv2 benchmarks demonstrate that DiffPixelFormer-L achieves mIoU scores of 54.28\% and 59.95\%, outperforming DFormer-L by 1.78\% and 2.75\%, respectively.
Code is available at \href{https://github.com/gongyan1/DiffPixelFormer}{\textcolor{blue}{https://github.com/gongyan1/DiffPixelFormer}}.

\end{abstract}


\section{Introduction}

\IEEEPARstart{I}{ndoor} semantic segmentation is crucial for applications such as autonomous navigation~\cite{seichter2021efficient},~\cite{zhang2023multi},~\cite{zhang2022openmpd},~\cite{gong2025progressive},~\cite{gong2025stable}, augmented reality~\cite{zhou2022pgdenet}, and smart homes~\cite{zheng2024indoor},~\cite{zhou2023utlnet},~\cite{gong2023sifdrivenet}.
RGB-based methods~\cite{cai2022rgb},~\cite{li2025construction},~\cite{wang2020multi} suffer from illumination and texture variations, while depth-based ones are affected by noise and lack appearance cues~\cite{zhou2022rfnet},~\cite{zhao2023cross},~\cite{chen2025dsdp}.
Consequently, leveraging the complementary strengths of RGB-D modalities thus becomes essential for improving robustness and accuracy in indoor semantic segmentation.

\begin{figure}[!t]
\centering
\includegraphics[width=\linewidth]{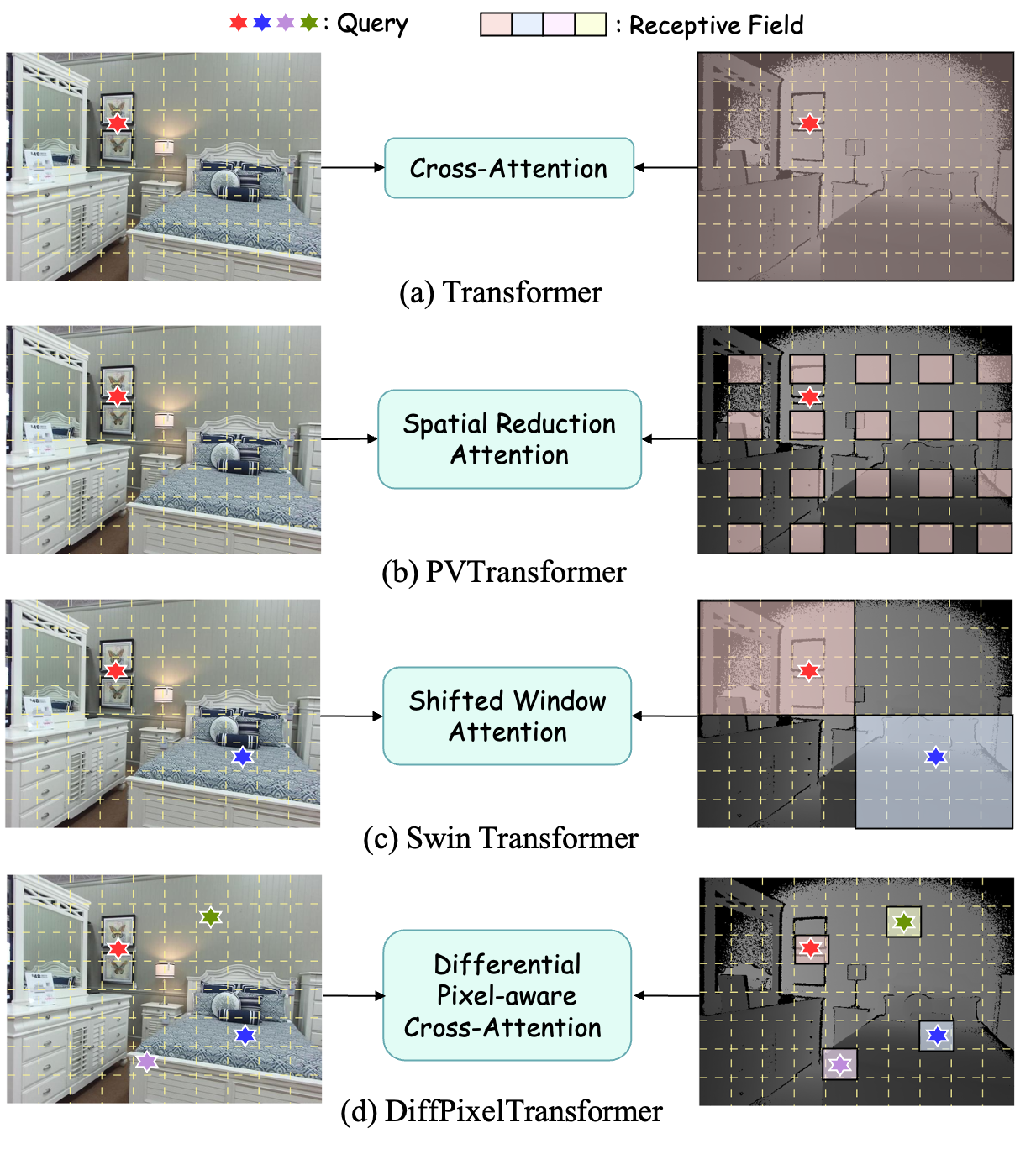}
\caption{Comparison of receptive fields among different cross-attention-based multimodal fusion methods.}
\label{fig_1}
\end{figure}

In multimodal learning, fusion strategies are broadly categorized into \textit{exchange-based} and \textit{interaction-based} paradigms. \textit{Exchange-based methods} emphasize efficiency by leveraging spatial correspondences for information substitution, e.g., Wang et al.~\cite{wang2020makes} with position mapping and TokenFusion~\cite{wang2022multimodal} using token substitution. However, such approaches often cause irreversible loss of modality-specific features, particularly in complementary modalities, thereby limiting performance. 
In contrast, \textit{interaction-based methods} explicitly model cross-modal dependencies to enhance representational capacity. Early works used simple concatenation~\cite{su2019vl}, while recent studies increasingly rely on attention mechanisms~\cite{zhou2022rfnet},~\cite{zhao2023cross},~\cite{zhou2023cmpffnet},~\cite{gong2024tclanenet},~\cite{gong2024steering}, with cross-attention (CA)~\cite{vaswani2017attention},~\cite{dong2022cswin} proving effective for capturing fine-grained relations. However, global dependency modeling incurs prohibitive complexity, as shown in Fig.~\ref{fig_1} (a). To address this, PVTransformer~\cite{leng2024pvtransformer} and Swin Transformer~\cite{liu2021swin} introduce sparse connections, local windows, or downsampled features, as shown in Fig.~\ref{fig_1} (b) and (c).
However, these methods remain computationally heavy and degrade pixel-level alignment, limiting local cross-modal correlation modeling in RGB-D data. To address this, we propose differential pixel-aware cross-attention (PACA), as illustrated in Fig.~\ref{fig_1} (d), which efficiently captures fine-grained, pixel-level cross-modal dependencies.

Through a systematic investigation of existing RGB-D fusion methods based on interaction-based paradigms, we identify two primary limitations. First, most approaches \cite{zhou2022pgdenet},~\cite{zheng2024indoor},~\cite{yin2025dformerv2},~\cite{yin2023dformer} fail to jointly consider intra- and inter-modal feature modeling: intra-modal representations lack sufficient capture of long-range dependencies and global context, limiting single-modality expressiveness, while inter-modal interactions often rely on coarse-grained cross-attention, which struggles to achieve pixel-level alignment and fine-grained correlation modeling. Second, existing methods \cite{seichter2021efficient},~\cite{zhao2023cross},~\cite{zhou2023cmpffnet},~\cite{gupta2014learning} do not jointly model shared and differential information across modalities. Although RGB and depth modalities exhibit common structural cues such as edges and contours, they also contain complementary features, including RGB texture and color and depth geometry. Current approaches~\cite{zhou2023cmpffnet},~\cite{jia2024geminifusion} typically ignore this distinction and perform undifferentiated aggregation, weakening the discriminative power and representational capacity of the fused features and constraining overall performance.

In this paper, we propose DiffPixelFormer, a Differential Pixel-aware Transformer for RGB-D indoor scene segmentation that simultaneously strengthens intra-modal representations and models inter-modal interactions. 
Specifically, the Intra-Inter Modal Interaction Block (IIMIB) explicitly separates intra- and inter-modal learning. Intra-modal interactions leverage self-attention (SA) to capture long-range dependencies and global context, enhancing the robustness and semantic completeness of single-modality features, while inter-modal interactions employ the Differential–Shared Inter-Modal (DSIM) module to disentangle modality-specific and shared information. Within DSIM, a difference discriminator ensures precise pixel-level alignment and models modality-specific discrepancies, whereas a similarity discriminator extracts shared structural cues such as edges and contours.
Furthermore, an adaptive fusion factor dynamically balances differential and shared cues to produce discriminative, semantically consistent cross-modal representations.
Extensive experiments on the NYUv2 and SUN RGB-D benchmarks demonstrate that DiffPixelFormer achieves significantly superior segmentation performance compared to state-of-the-art methods. Furthermore, as illustrated in Fig. \ref{fig_2}, DiffPixelFormer markedly reduces parameter count relative to representative cross-attention schemes, thereby attaining a favorable balance between efficiency and accuracy.
The contribution of this article can be summarized as follows:

\begin{figure}[!t]
\centering
\includegraphics[width=1.0 \linewidth]{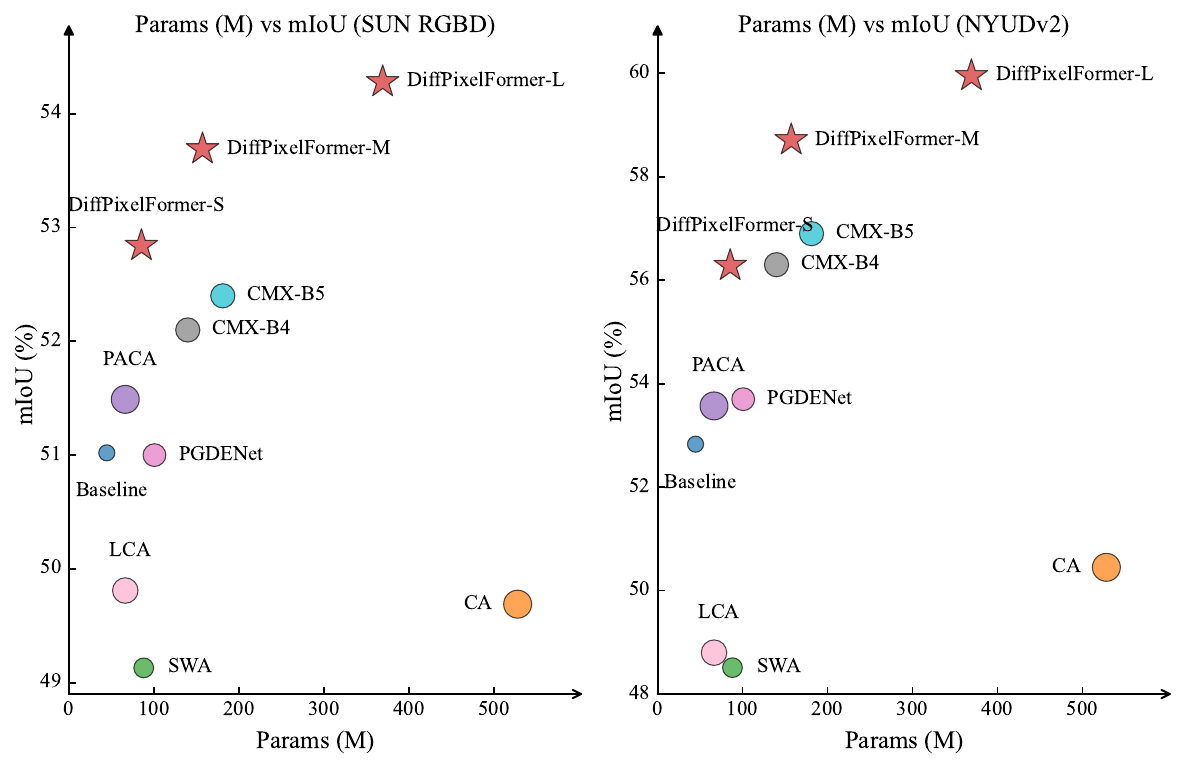}
\caption{Performance comparison of different attention variants on SUN RGB-D and NYUDv2 in terms of parameters and mIoU. “CA” denotes Cross-Attention, “SWA” denotes Shifted Window Attention, “LCA” denotes Local Cross-Attention, and ``PACA'' denotes Pixel-Aware Cross-Attention.}
\label{fig_2}
\end{figure}

\begin{itemize}
\item We propose DiffPixelFormer, a novel Differential Pixel-aware Transformer for RGB-D indoor scene segmentation, whose core IIMIB jointly models intra-modal and inter-modal interactions to fully exploit complementary and shared information.

\item We introduce DSIM as the inter-modal component of IIMIB, which disentangles modality-specific and shared cues to achieve fine-grained, pixel-level alignment and semantically consistent cross-modal representations.

\item Extensive experiments on NYUv2 and SUN RGB-D benchmarks demonstrate that DiffPixelFormer consistently outperforms state-of-the-art methods in segmentation accuracy while achieving a favorable trade-off between performance and efficiency.

\end{itemize}

\section{Related Work}
\subsection{RGB-D Indoor Semantic Segmentation}


Indoor semantic segmentation is a fundamental task in computer vision and robotic perception, supporting applications such as autonomous navigation, augmented reality, and smart homes. RGB-based methods, including FCN~\cite{long2015fully}, U-Net~\cite{ronneberger2015u}, TokenFusion-B5~\cite{wang2022multimodal}, and DFormerV2~\cite{yin2025dformerv2}, have achieved remarkable progress but remain vulnerable to illumination changes, texture degradation, occlusion, and noise. To improve robustness under weak lighting and motion blur, recent approaches integrate depth cues, which provide geometric priors complementary to appearance features~\cite{gupta2014learning,silberman2012indoor}. Building on this, DFormer~\cite{yin2023dformer} introduced large-scale RGB-D pretraining on ImageNet-1K, while DFormerV2~\cite{yin2025dformerv2} treats depth as implicit geometric priors. PDDM~\cite{xu2025pddm} alleviates depth scarcity via pseudo-depth generation, and EACNet~\cite{mao2025eacnet} designs lightweight fusion for resource-constrained platforms. CMX~\cite{zhang2023cmx} enables multi-scale cross-modal interaction, ACNet~\cite{hu2019acnet} applies attention-based adaptive fusion, and SegFormer~\cite{xie2021segformer} demonstrates Transformer-based unified representation learning.

\textit{Difference:}
Existing methods rely on coarse cross-attention, which limits pixel-level alignment and incurs substantial computational overhead.
DiffPixelFormer enhances intra-modal features and models inter-modal relations via pixel-aware differential and similarity attention, achieving precise and efficient representations.

\subsection{Multimodal Fusion Strategies}

Multimodal fusion in RGB-D segmentation harnesses the complementary strengths of RGB images, which provide rich appearance and semantic information, and depth maps, which offer stable geometric priors. Existing research has mainly explored three categories of fusion strategies.
(1) Interaction-based fusion models cross-modal dependencies to exploit complementarity, as in ACNet~\cite{hu2019acnet}, RFNet~\cite{zou2022rgb}, and Primkd~\cite{hao2024primkd}, which employ cross-modal attention to dynamically weight features. However, the global modeling nature of cross-modal attention necessitates dense pairwise interactions between modalities, substantially inflating computational and memory costs, and thus hindering real-time deployment in practical scenarios.
(2) Alignment-based fusion enforces feature-level consistency to alleviate distributional gaps and noise. For instance, GeminiFusion~\cite{jia2024geminifusion} integrates intra- and inter-modal attention with noise control, while CMX~\cite{zhang2023cmx} scales fusion to RGB, depth, and LiDAR. Such approaches often neglect differential modality cues and impose excessive constraints on feature interactions, thereby limiting the exploitation of modality-specific representations.
(3) Backbone-enhanced fusion introduces adaptive modules or dual-path designs to strengthen integration efficiency and representation robustness~\cite{xie2021segformer},\cite{yu2018bisenet},\cite{yuan2023pointmbf}.
These architectures frequently depend on elaborate network designs, which not only elevate implementation and optimization complexity, but also impede scalability and adaptability across heterogeneous multimodal scenarios.

\textit{Difference:}
Existing fusion strategies neglect the distinction between intra-modal and inter-modal interactions and fail to separate shared from modality-specific information. DiffPixelFormer addresses these limitations by combining intra-modal self-attention with a pixel-wise DSIM module, explicitly disentangling shared and differential cues to enable fine-grained, discriminative cross-modal representations.

\section{Method}

\begin{figure*}[!t]
\centering
\includegraphics[width=1.0 \linewidth]{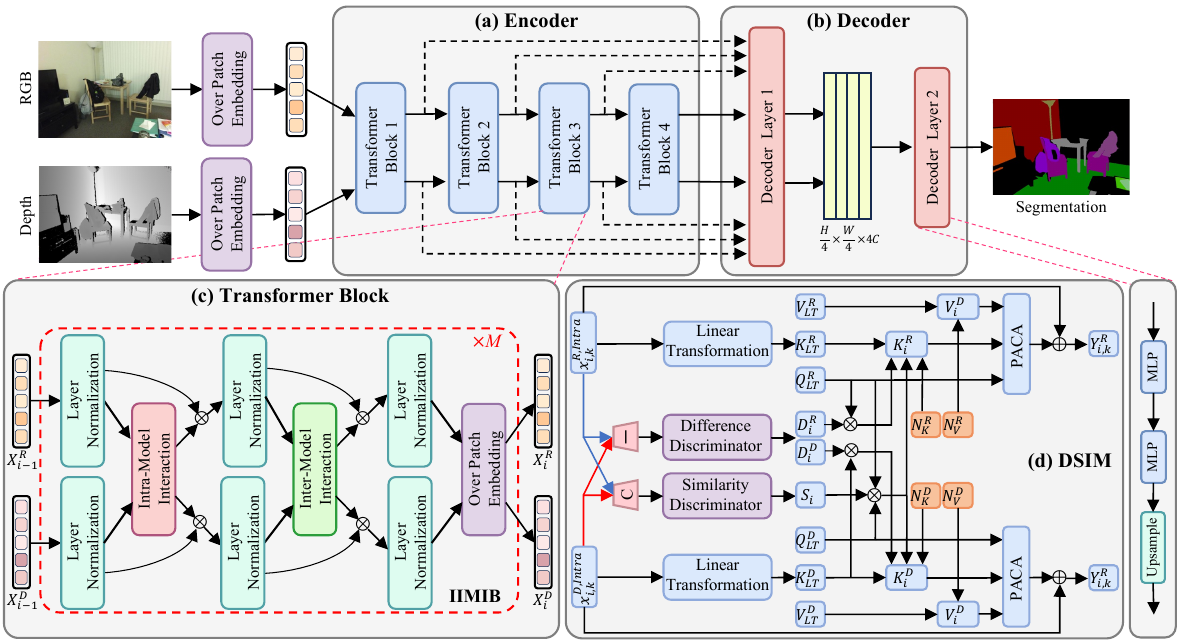}
\caption{The overall architecture of DiffPixelFormer adopts an encoder–decoder design, where the encoder employs multiple Intra-Inter Modal Interaction Blocks (IIMIBs) for efficient intra- and inter-modal fusion, and the decoder restores spatial and semantic details via multi-scale aggregation.}
\label{overall-network}
\end{figure*}

In this section, we first review commonly used cross-attention mechanisms in Section \ref{cross-attention}. We then present the overall architecture of DiffPixelFormer in Section \ref{overall-architecture}, followed by a detailed discussion of intra- and inter-modal feature interactions in Section \ref{intra-inter model interaction}.

\subsection{Cross-Attention Review}
\label{cross-attention}
Cross-Attention (CA) is widely employed in multimodal fusion to capture inter-modal dependencies by computing attention weights between the queries of one modality and the key-value pairs of another, thereby enabling effective information transfer and integration. For $N$ patches, let the RGB and depth feature maps be $\mathbf{X}^R, \mathbf{X}^D \in \mathbb{R}^{N \times d}$, where $d$ is the feature dimension, and the cross-attention computation can then be defined as follows:
\begin{equation}
\begin{aligned}
& \mathbf{Y}^R=\operatorname{CA}\left(\mathbf{X}^R \mathrm{~W}^{\mathrm{Q}}, \mathbf{X}^D \mathrm{~W}^{\mathrm{K}}, \mathbf{X}^D \mathrm{~W}^{\mathrm{V}}\right)+\mathbf{X}^R \\
& \mathbf{Y}^D=\operatorname{CA}\left(\mathbf{X}^D \mathrm{~W}^{\mathrm{Q}}, \mathbf{X}^R \mathrm{~W}^{\mathrm{K}}, \mathbf{X}^R \mathrm{~W}^{\mathrm{V}}\right)+\mathbf{X}^D \\
& \text{CA}(\mathrm{Q}, \mathrm{K}, \mathrm{V})=\operatorname{Softmax}\left(\frac{\mathrm{QK}^{\mathrm{T}}}{\sqrt{d}}\right)\mathrm{V}
\end{aligned}
\end{equation}
where $\mathrm{W}^{Q}, \mathrm{W}^{K}, \mathrm{W}^{V}$ are learnable projection matrices for queries ($\mathrm{Q}$), keys ($\mathrm{K}$), and values ($\mathrm{V}$), 
$\mathbf{Y}^R$ and $\mathbf{Y}^D$ denote the updated RGB and depth features.
The computational complexity of the above operation is $O(N^2 d)$, where the quadratic growth imposes a heavy burden when handling large-scale data or high-resolution images. Furthermore, cross-attention mechanisms typically rely on coarse-grained feature aggregation, which hinders pixel-level alignment and fine-grained correspondence modeling, thereby limiting the granularity and expressiveness of multimodal fusion.

\subsection{Network Overview}
\label{overall-architecture}

DiffPixelFormer adopts a mainstream encoder-decoder architecture, as illustrated in Fig.~\ref{overall-network}. Initially, RGB and depth images are converted into encoded features $X^{R}_0$ and $X^{D}_0$ via overlapped patch embedding. These encoded features are then sequentially fed into four Transformer Blocks, progressively extracting and generating multi-resolution feature maps, denoted as $X^{R}_i$ and $X^{D}_i$ for $i=1,2,3,4$. For each Transformer Block, the input features $X^{R}_{i-1}$ and $X^{D}_{i-1}$ are processed by $M$ Intra-Inter Modal Interaction Blocks (IIMIBs) to produce the output features $X^{R}_{i}$ and $X^{D}_{i}$, which can be formulated as:
\begin{equation}
(X^{R}_i, X^{D}_i) = \mathcal{F}_M \circ \mathcal{F}_{M-1} \circ \cdots \circ \mathcal{F}_1 (X^{R}_{i-1}, X^{D}_{i-1}),
\end{equation}
where $\mathcal{F}_M$ denotes the operation of the $M$-th IIMIB. In practice, the four Transformer Blocks employ 3, 6, 4, and 3 IIMIBs, respectively.


Each IIMIB applies modality-specific layer normalization, followed by intra- and inter-modal interaction modules that respectively capture long-range dependencies and jointly model shared and specific representations. Subsequent normalization and overlapped patch merging enable spatial aggregation and dimensionality reduction, while residual connections preserve information flow and mitigate feature degradation.
Notably, layer normalization is applied independently to each modality to account for distributional disparities, whereas the remaining layers share parameters to improve computational efficiency.
The decoder fuses multi-scale features and refines them through upsampling and dual MLP layers to generate the final segmentation map.

\subsection{Intra-Inter Modal Interaction Block (IIMIB)}
\label{intra-inter model interaction}

Due to substantial differences in data distribution, semantic structure, and noise across modalities, direct fusion may amplify noise or obscure useful information. To address this, we introduce the IIMIB module (see Fig.~\ref{overall-network} (c)), which explicitly separates intra-modal and inter-modal interactions to better exploit multimodal features. 
Intra-modal interaction captures long-range dependencies and enhances feature discriminability within each modality, while inter-modal interaction integrates complementary information across modalities.
This hierarchical strategy enables more effective multimodal collaboration, as detailed below.

\subsubsection{Intra-Modal Interaction}
To fully exploit the intrinsic characteristics of each modality and enhance the quality of inputs for inter-modal fusion, we employ an intra-modal interaction module based on self-attention (SA) for both RGB and depth modalities $X^{R}_{i}$ and $X^{D}_{i}$, defined as follows:
\begin{equation}
(X^{R,\text{Intra}}_i,\, X^{D,\text{Intra}}_i) = SA(X^{R}_{i},\, X^{D}_{i}),
\end{equation}
where $X^{R,\text{Intra}}_i$ and $X^{D,\text{Intra}}_i$ denote the outputs with enhanced features after intra-modal interaction.
By adaptively attending to salient information and modeling long-range dependencies within each modality, more discriminative and comprehensive intra-modal representations are obtained, serving as high-quality inputs for subsequent multimodal fusion.

\subsubsection{Inter-Modal Interaction}

Existing cross-attention-based multimodal fusion methods exploit cross-modal correlations but often neglect complementary modality-specific cues and suffer from heavy computational and parameter overhead due to dense token interactions, many of which are uninformative. To address these issues, we argue that inter-modal fusion should focus on corresponding spatial locations, emphasizing discriminative and differential modality information. Accordingly, we propose DSIM (see Fig.~\ref{overall-network} (d)), which adaptively extracts fine-grained modality-specific features and computes cross-modal similarity for shared representation.

Firstly, we design two lightweight relation discriminators: a difference discriminator and a similarity discriminator. Given the input features $X^{R,\text{Intra}}_i$ and $X^{D,\text{Intra}}_i$, the normalized difference and similarity scores are computed as follows:
\begin{equation}
\left\{
\begin{aligned}
&\mathcal{D}^R_i = f^R_d(X^{R,\text{Intra}}_i - X^{D,\text{Intra}}_i) \\
&\mathcal{D}^D_i = f^D_d(X^{D,\text{Intra}}_i - X^{R,\text{Intra}}_i) \\
&\mathcal{S}_i   = f_s([X^{R,\text{Intra}}_i,\, X^{D,\text{Intra}}_i])
\end{aligned}
\right.
\end{equation}
where $\mathcal{D}^R_i, \mathcal{D}^D_i, \mathcal{S}_i \in [0, 1]$ quantify the modality-specific distinction for RGB and Depth, and the shared similarity, respectively.
$[.]$ denotes the concatenation operation.
Here, $f_d(\cdot)$ and $f_s(\cdot)$ denote the difference and similarity discriminators, both implemented as two-layer MLPs with a softmax activation. Note that, due to the inherent asymmetry of inter-modal feature differences, bidirectional modeling is required to fully capture complementary information, whereas the similarity score is uniquely defined for each feature pair without the need for directional distinction.

Subsequently, features from different modalities are projected via linear transformation $LT$ to generate the $Q_{LT}$, $K_{LT}$, and $V_{LT}$. In standard cross-attention, there is a tendency to disproportionately learn from another modality, resulting in over-reliance on similar components and neglect of less similar yet potentially important information. To mitigate this, we modulate the keys of each modality using the previously computed $\mathcal{D}^R_i$, $\mathcal{D}^D_i$, and $\mathcal{S}_i$ scores, and replace the keys in Eq~(1) as follows:
\begin{equation}
\begin{aligned}
K^R_i &= [\alpha^R * Q^R_{LT} \ast \mathcal{D}^R_i,\, \beta^R * Q^R_{LT} \ast \mathcal{S}_i], \\
K^D_i &= [\alpha^D * Q^D_{LT} \ast \mathcal{D}^D_i,\, \beta^D * Q^D_{LT} \ast \mathcal{S}_i],
\end{aligned}
\end{equation}
where $\alpha$ and $\beta$ denote learnable factors.
Furthermore, we construct the embeddings $V^R_i$ and $V^D_i$ by combining the differential feature representations with the complementary modality features, formulated as:
\begin{equation}
V^R_i = [V^{R}_{LT} - V^{D}_{LT}, \; V^{D}_{LT}], \quad
V^D_i = [V^{D}_{LT} - V^{R}_{LT}, \; V^{R}_{LT}],
\end{equation}
where the ordering of $V^R_i$ and $V^D_i$ is kept consistent with $K^R_i$ and $K^D_i$ to facilitate the computation of pixel-aware cross-attention.

However, deriving both $Q$ and $K$ from the same modality introduces self-referential bias in the attention scores, which overemphasizes intra-modal correlations at the expense of complementary cross-modal cues, ultimately limiting the effectiveness of multimodal fusion.
To address this limitation, adaptive noise terms $N^R$ and $N^D$ are injected at different layers, formulated as:
\begin{equation}
\begin{aligned}
Q^{R'}_i &= \mathbf{X}^R_{i} W^{Q}, & Q^{D'}_i &= \mathbf{X}^D_{i} W^{Q}, \\
K^{R'}_i &= [N^R_K, K^R_i] W^{K}, & K^{D'}_i &= [N^D_K, K^D_i] W^{K}, \\
V^{R'}_i &= [N^R_V, V^R_i] W^{V}, & V^{D'}_i &= [N^D_V, V^D_i] W^{V},
\end{aligned}
\end{equation}
thereby increasing the diversity of feature representations and enabling more effective cross-modal integration. The outputs $\mathbf{Y}^R_{i}$ and $\mathbf{Y}^D_{i}$ are obtained via pixel-aware cross-attention, defined as:
\begin{equation}
\begin{aligned}
\mathbf{Y}^R_{i, k} &= \operatorname{CA}(Q^{R'}_{i, k}, K^{D'}_{i, k}, V^{D'}_{i, k}) + \mathbf{X}^{R,Intra}_{i, k},\\
\mathbf{Y}^D_{i, k} &= \operatorname{CA}(Q^{D'}_{i, k}, K^{R'}_{i, k}, V^{R'}_{i, k}) + \mathbf{X}^{D,Intra}_{i, k},
\end{aligned}
\end{equation}
where $k$ denotes the $k$-th token. 
Notably, cross-attention is restricted to tokens at the same spatial location rather than across all tokens, thereby preserving spatial correspondence while enabling fine-grained multimodal interaction.

\section{Experiment}
This section presents a comprehensive evaluation of DiffPixelFormer. Sections~\ref{IV-A} and~\ref{IV-B} detail the experimental setup and implementation. Section~\ref{IV-C} compares our method with state-of-the-art approaches, followed by quantitative analysis in Section~\ref{IV-D}. Ablation studies in Section~\ref{IV-E} examine the impact of fusion strategies, self-attention mechanisms, relation discriminators, and backbone on performance.

\subsection{Datasets and Evaluation Settings} \label{IV-A}
\noindent
\textbf{SUN RGB-D:} SUN RGB-D~\cite{song2015sun} is a large-scale RGB-D dataset designed for indoor scene understanding, containing 10,335 images with pixel-level semantic annotations. It provides a standard split of 5,285 training and 5,050 testing samples, and incorporates data from NYU Depth V2~\cite{silberman2012indoor} and Berkeley B3DO~\cite{janoch2011category}. The dataset encompasses a broad spectrum of indoor environments, including bedrooms, living rooms, offices, kitchens, and classrooms, serving as a comprehensive benchmark for scene understanding research.

\noindent
\textbf{NYUDv2:} NYUDv2~\cite{silberman2012indoor} is a widely-used RGB-D dataset for indoor scene understanding, comprising 1,449 densely annotated images with corresponding depth maps, captured using a Microsoft Kinect sensor across 464 distinct indoor scenes. Each annotated image provides both class and instance-level segmentation. Additionally, the dataset includes 407,024 unlabeled frames. NYUDv2 covers a broad spectrum of indoor environments, such as dining rooms, living rooms, bedrooms, bathrooms, and offices.

\noindent
\textbf{Evaluation metrics:} 
To quantitatively assess the segmentation performance, we employ three widely used evaluation metrics, namely \textit{mean Intersection over Union} (mIoU), \textit{mean Pixel Accuracy} (mAcc), and \textit{Pixel Accuracy} (Pixel Acc). Their mathematical definitions are given as follows:
\begin{equation}
\text{mIoU} = \frac{1}{k} \sum_{i=1}^{k} \frac{p_{ii}}{\sum_{j=1}^{k} p_{ij} + \sum_{j=1}^{k} p_{ji} - p_{ii}},
\label{eq:miou}
\end{equation}
\begin{equation}
\text{mAcc} = \frac{1}{k} \sum_{i=1}^{k} \frac{p_{ii}}{\sum_{j=1}^{k} p_{ij}},
\label{eq:macc}
\end{equation}
\begin{equation}
\text{Pixel Acc} = \frac{\sum_{i=1}^{k} p_{ii}}{\sum_{i=1}^{k} \sum_{j=1}^{k} p_{ij}}.
\label{eq:pa}
\end{equation}
Here, $k$ denotes the total number of classes, and $p_{ij}$ represents the number of pixels with ground-truth label $i$ that are predicted as class $j$.
Specifically, mIoU quantifies the mean intersection over union between predicted and ground truth regions across all classes, mAcc represents the average pixel-wise accuracy computed per class, and Pixel ACC denotes the overall proportion of correctly classified pixels in the dataset. Among these metrics, mIoU is regarded as the primary indicator for evaluating segmentation performance.

\subsection{Experiment setting and training details} \label{IV-B}
In multimodal semantic segmentation, our training settings largely follow the TokenFusion \cite{wang2022multimodal}. Experiments are conducted on NVIDIA V100 GPUs for both the NYUDv2 and SUN RGB-D datasets, under the same environmental conditions as the original works. The encoder backbone is adapted from SegFormer~\cite{xie2021segformer}, pre-trained only on ImageNet-1K. For both datasets, we adopt the training protocols of TokenFusion, ensuring consistency in batch size, optimizer, learning rate scheduler, and other key hyperparameters. In our DiffPixelFormer, the number of attention heads is fixed at 8, with a drop path rate of 0.4 and a drop rate of 0.0 to mitigate overfitting.
Except for setting the learning rate to $2 \times 10^{-4}$, all other hyperparameters, including batch size, optimizer, and weight decay, follow those of TokenFusion \cite{wang2022multimodal}.

\subsection{Comparison with SOTA Methods} \label{IV-C}

\begin{table*}[!t]
\centering
\caption{Comparison of existing state-of-the-art methods on the SUN RGB-D and NYUDv2 datasets, where \textbf{bold} and \underline{underlined} values indicate the best and second-best performances, and ``--'' denotes results not reported in the original papers.}
\renewcommand{\arraystretch}{1.05}
\setlength{\tabcolsep}{2.5pt}
\resizebox{\textwidth}{!}{
\begin{tabular}{l l c c c c c c c c c}
\hline
\multirow{2}{*}{\textbf{Model}} & \multirow{2}{*}{\textbf{Backbone}} & \multirow{2}{*}{\textbf{Publication}} & \multirow{2}{*}{\textbf{Year}} & \multirow{2}{*}{\textbf{Param(M)}} & 
\multicolumn{3}{c}{\textbf{SUN RGB-D}} & \multicolumn{3}{c}{\textbf{NYUDv2}} \\
\cline{6-11}
& & & & & \textbf{mIoU} & \textbf{Pixel Acc} & \textbf{mAcc} & \textbf{mIoU} & \textbf{Pixel Acc} & \textbf{mAcc} \\
\hline
SGNet~\cite{chen2021spatial} & ResNet-101 &  TIP & 2021 & 64.70 & \cellcolor{gray!15}48.60 & - & - & \cellcolor{gray!15}51.10 & - & - \\
ESANet~\cite{seichter2021efficient} & ResNet-34 & ICRA & 2021 & 31.20 & \cellcolor{gray!15}48.20 & - & - & \cellcolor{gray!15}50.30 & - & - \\
ShapeConv~\cite{cao2021shapeconv} & ResNeXt-101 & ICCV & 2021 & 86.80 & \cellcolor{gray!15}48.60 & - & - & \cellcolor{gray!15}51.30 & - & - \\
CEN~\cite{wang2022learning} & ResNet-50 &  TIP & 2022 & - & \cellcolor{gray!15}51.10 & \underline{83.50} & 63.20 & \cellcolor{gray!15}52.50 & 77.70 & 65.00 \\
MultiMAE~\cite{bachmann2022multimae} & ViT-Base & ECCV & 2022 & 95.20 & \cellcolor{gray!15}51.10 & - & - & \cellcolor{gray!15}56.00 & - & - \\
Omnivore~\cite{girdhar2022omnivore} & Swin-Small & CVPR & 2022 & 95.70 & \cellcolor{gray!15}- & - & - & \cellcolor{gray!15}54.00 & - & - \\
PGDENet~\cite{zhou2022pgdenet} & ResNet-34 &  TMM & 2022 & 100.70 & \cellcolor{gray!15}51.00 & - & - & \cellcolor{gray!15}53.70 & - & - \\
EMSANet~\cite{seichter2022efficient} & ResNet-34 & IJCNN & 2022 & 46.90 & \cellcolor{gray!15}50.90 & - & - & \cellcolor{gray!15}59.00 & - & - \\
TokenFusion-B2~\cite{wang2022multimodal} & MiT-B2 & CVPR & 2022 & 26.00 & \cellcolor{gray!15}50.30 & - & - & \cellcolor{gray!15}53.30 & - & - \\
TokenFusion-B3~\cite{wang2022multimodal} & MiT-B3 & CVPR & 2022 & 45.90 & \cellcolor{gray!15}51.40 & 82.80 & 63.60 & \cellcolor{gray!15}54.20 & 79.00 & 66.90 \\
TokenFusion-B5~\cite{wang2022multimodal} & MiT-B5 & CVPR & 2022 & 83.30 & \cellcolor{gray!15}51.80 & 83.10 & 63.90 & \cellcolor{gray!15}55.10 & 79.10 & 67.50 \\
CMX-B2~\cite{zhang2023cmx} & MiT-B2 &  TITS & 2023 & 66.60 & \cellcolor{gray!15}49.70 & - & - & \cellcolor{gray!15}54.40 & - & - \\
CMX-B4~\cite{zhang2023cmx} & MiT-B4 &  TITS & 2023 & 139.90 & \cellcolor{gray!15}52.10 & - & - & \cellcolor{gray!15}56.30 & - & - \\
CMX-B5~\cite{zhang2023cmx} & MiT-B5 &  TITS & 2023 & 181.10 & \cellcolor{gray!15}52.40 & - & - & \cellcolor{gray!15}56.90 & - & - \\
CMNext~\cite{zhang2023delivering} & MiT-B4 & CVPR & 2023 & 119.60 & \cellcolor{gray!15}51.90 & - & - & \cellcolor{gray!15}56.90 & - & - \\
PGDENet~\cite{zhou2022pgdenet} & UNet & TMM & 2023 & - & \cellcolor{gray!15}51.00 & - & 61.70 & \cellcolor{gray!15}53.70 & - & 66.70 \\
DPLNet~\cite{dong2024efficient} & MiT-B5 & IROS & 2024 & - & \cellcolor{gray!15}52.80 & - & - & \cellcolor{gray!15}58.30 & - & - \\
GeminiFusion~\cite{jia2024geminifusion} & MiT-B3 & ICML & 2024 & 75.80 & \cellcolor{gray!15}52.70 & 83.30 & 64.60 & \cellcolor{gray!15}56.80 & 79.90 & 69.90 \\
DFormer-T~\cite{yin2023dformer} & DFormer-Tiny & ICRL & 2024 & 6.00 & \cellcolor{gray!15}48.80 & - & - & \cellcolor{gray!15}51.80 & - & - \\
DFormer-S~\cite{yin2023dformer} & DFormer-Small & ICRL & 2024 & 18.70 & \cellcolor{gray!15}50.00 & - & - & \cellcolor{gray!15}53.60 & - & - \\
DFormer-B~\cite{yin2023dformer} & DFormer-Base & ICRL & 2024 & 29.50 & \cellcolor{gray!15}51.20 & - & - & \cellcolor{gray!15}55.60 & - & - \\
DFormer-L~\cite{yin2023dformer} & DFormer-Large & ICRL & 2024 & 39.00 & \cellcolor{gray!15}52.50 & - & - & \cellcolor{gray!15}57.20 & - & - \\
AsymFormer~\cite{du2024asymformer} & MiT-B0+ConvNeXt-Tin & CVPR & 2024 & 33.00 & \cellcolor{gray!15}49.10 & - & - & \cellcolor{gray!15}55.30 & - & - \\
PolyMaX~\cite{yang2024polymax} & ConvNeXt-L & WACV & 2024 & - &\cellcolor{gray!15} - & - & - & \cellcolor{gray!15}58.10 & - & - \\
DFormerV2-S~\cite{yin2025dformerv2} & DFormerV2-Small & CVPR & 2025 & 26.70 & \cellcolor{gray!15}51.50 & - & - & \cellcolor{gray!15}56.00 & - & - \\
DFormerV2-B~\cite{yin2025dformerv2} & DFormerV2-Base & CVPR & 2025 & 53.90 & \cellcolor{gray!15}52.80 & - & - & \cellcolor{gray!15}57.70 & - & - \\
DFormerV2-L~\cite{yin2025dformerv2} & DFormerV2-Large & CVPR & 2025 & 95.50 & \cellcolor{gray!15}53.30 & - & - & \cellcolor{gray!15}58.40 & - & - \\
EACNet~\cite{mao2025eacnet} & ConvNeXt-T+VAN-B0 & DDCLS & 2025 & 37.00 & \cellcolor{gray!15}52.60 & - & - & \cellcolor{gray!15}57.60 & - & - \\
FCDENet~\cite{zhou2025feature} & MiT-B4 & IoT & 2025 & 128.90 & \cellcolor{gray!15}52.00 & 82.60 & - & \cellcolor{gray!15}57.70 & 80.00 & - \\
DCANet~\cite{bai2025dcanet} & VMamba & PR & 2025 & 123.80 & \cellcolor{gray!15}49.60 & 82.60 & - & \cellcolor{gray!15}53.30 & 78.20 & - \\
Sigma~\cite{wan2025sigma} & VMamba & WACV & 2025 & 69.80 & \cellcolor{gray!15}52.40 & - & - & \cellcolor{gray!15}57.00 & - & - \\
ADBNet~\cite{xu2025adbnet} & ConvNeXt & KBS & 2025 & 45.90 & \cellcolor{gray!15}49.60 & 82.30 & - & \cellcolor{gray!15}56.00 & 79.40 & - \\
ECMRN~\cite{jia2025ecmrn} & DFormer & KBS & 2025 & 68.60 & \cellcolor{gray!15}52.90 & - & - & \cellcolor{gray!15}58.10 & - & - \\
DFNet-L~\cite{yang2025difference} & MiT-B4 & TII & 2025 & 108.75 & \cellcolor{gray!15}51.73 & 83.42 & - & \cellcolor{gray!15}57.33 & 79.88 & - \\
Sigma~\cite{wan2025sigma} & VMamba & WACV & 2025 & 69.80 & \cellcolor{gray!15}52.40 & - & - & \cellcolor{gray!15}57.00 & - & - \\
\rowcolor{gray!20}
DiffPixelFormer-S (ours) & MiT-B3 & - & 2025 & 85.41 & 52.84 & 83.37 & 64.83 & 56.28 & 79.79 & 68.98 \\
\rowcolor{gray!20}
DiffPixelFormer-M (ours) & MiT-B5 & - & 2025 & 157.24 & \underline{53.69} & 83.47 & \textbf{66.11} & \underline{58.71} & \underline{80.30} & \underline{69.93} \\
\rowcolor{gray!20}
DiffPixelFormer-L (ours) & Swin-Large & - & 2025 & 369.22 & \textbf{54.28} & \textbf{84.14} & \underline{65.91} & \textbf{59.95} & \textbf{81.70} & \textbf{72.74} \\
\hline
\end{tabular}
}
\label{tab:sun-nyu}
\end{table*}

To thoroughly evaluate the effectiveness and superiority of the proposed DiffPixelFormer, we conduct extensive comparative experiments on two widely used RGB-D semantic segmentation benchmarks, SUN RGB-D and NYUDv2, as shown in Table~\ref{tab:sun-nyu}. The results demonstrate that DiffPixelFormer consistently outperforms state-of-the-art methods, achieving the best performance with mIoU scores of \textbf{54.28\%} on SUN RGB-D and \textbf{59.95\%} on NYUDv2. A more detailed analysis of the experiment results is presented below.

\subsubsection{Results on SUN RGB-D}
As shown in Table~\ref{tab:sun-nyu}, DiffPixelFormer-S achieves \textbf{52.84\%} mIoU on the SUN RGB-D dataset, clearly surpassing MultiMAE~\cite{bachmann2022multimae}, ShapeConv~\cite{cao2021shapeconv}, TokenFusion~\cite{wang2022multimodal}, and GeminiFusion~\cite{jia2024geminifusion} under comparable parameter budgets, highlighting superior parameter efficiency. With deeper backbones, DiffPixelFormer-M and DiffPixelFormer-L further improve performance to \textbf{53.69\%} and \textbf{54.28\%} mIoU, gains of 0.85\% and 1.44\% over DiffPixelFormer-S, confirming strong scalability. Under the same MiT-B5 backbone \cite{xie2021segformer}, DiffPixelFormer-M consistently outperforms TokenFusion-B5~\cite{wang2022multimodal}, CMX-B5~\cite{zhang2023cmx}, and DPLNet~\cite{dong2024efficient}, underscoring robustness and competitiveness across methods. Moreover, DiffPixelFormer-L exhibits strong competitiveness in large-scale settings, reaching 54.28\%, 84.14\%, and 65.91\% on mIoU, Pixel ACC, and mAcc, respectively. Overall, DiffPixelFormer demonstrates clear advantages across parameter scales and baselines, striking a favorable balance between complexity and generalization, and thus offering substantial practical value and research significance.

\subsubsection{Results on NYUDv2}
As shown in Table~\ref{tab:sun-nyu}, DiffPixelFormer demonstrates strong competitiveness across multiple configurations. The lightweight DiffPixelFormer-S achieves \textbf{56.28\%} mIoU with a moderate parameter scale, outperforming methods such as MultiMAE~\cite{bachmann2022multimae}, CMX-B4~\cite{zhang2023cmx}, DFormer-S~\cite{yin2023dformer}, DFormerV2-S~\cite{yin2025dformerv2}, and AsymFormer~\cite{du2024asymformer}, thereby establishing a clear performance advantage. Furthermore, under the same MiT-B5 backbone~\cite{xie2021segformer}, DiffPixelFormer-M attains \textbf{58.71\%} mIoU, consistently surpassing TokenFusion-B5~\cite{wang2022multimodal}, CMX-B5~\cite{zhang2023cmx}, and DPLNet~\cite{dong2024efficient}, highlighting its strong competitiveness under identical backbone settings. In the large-scale setting, DiffPixelFormer-L further improves performance to \textbf{59.95\%} mIoU, while achieving \textbf{81.70\%} Pixel Accuracy and \textbf{72.74\%} mAcc, outperforming the strongest existing models and demonstrating clear advantages for high-precision applications. Overall, DiffPixelFormer delivers stable and superior performance across different parameter scales and backbone choices, while maintaining flexibility to adapt to diverse architectures, thus exhibiting strong generalization ability and practical utility in complex indoor scene understanding.


\subsection{Quantitative Results}\label{IV-D}

\begin{figure*}[!t]
\centering
\includegraphics[width=1.0 \linewidth]{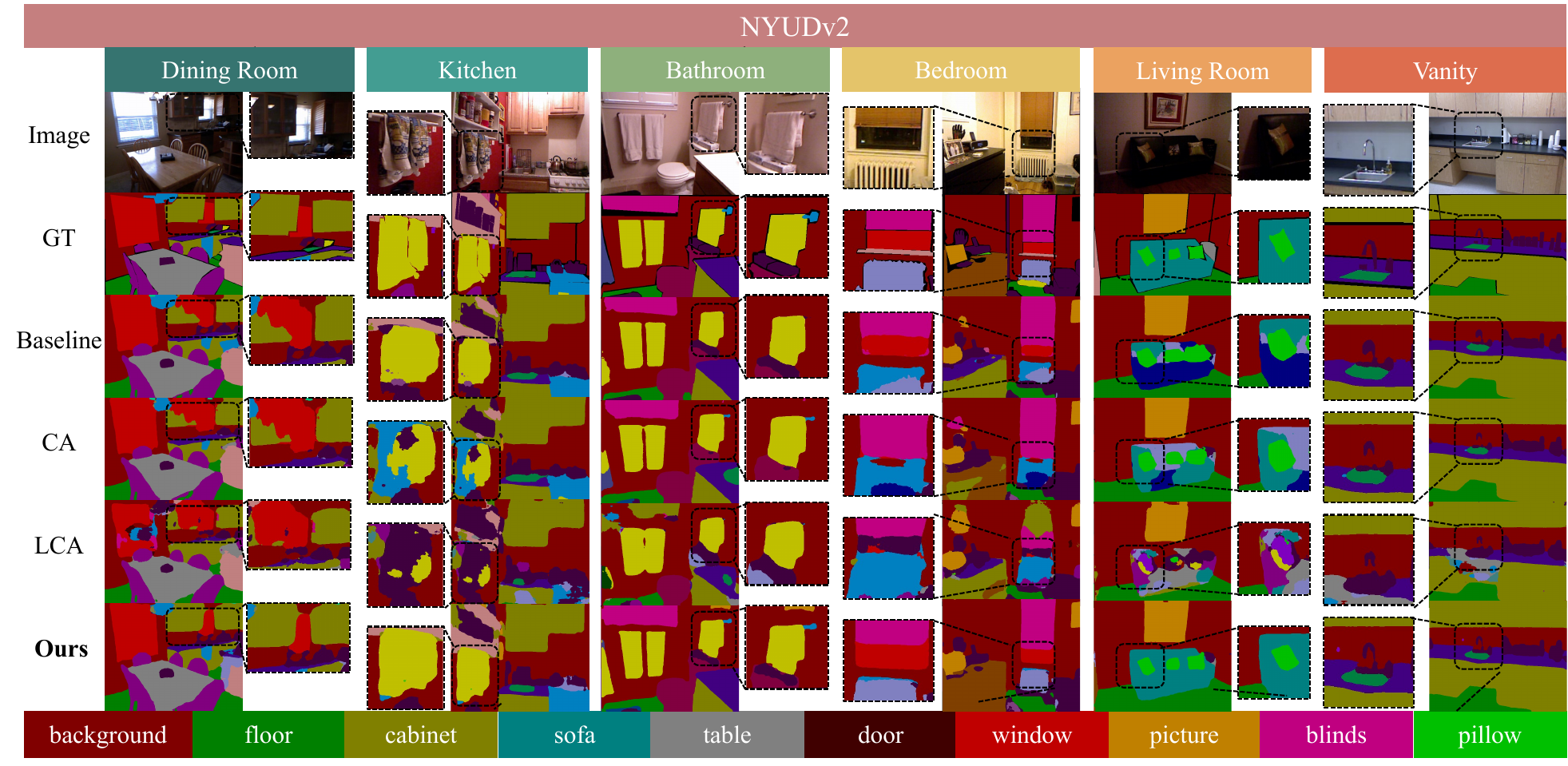}
\caption{Quantitative comparison of our DiffPixelFormer with the baseline and various cross-attention methods on NYUDv2, where GT denotes the ground truth.}
\label{results_nyudv2}
\end{figure*}

\begin{figure*}[!ht]
\centering
\includegraphics[width=1.0 \linewidth]{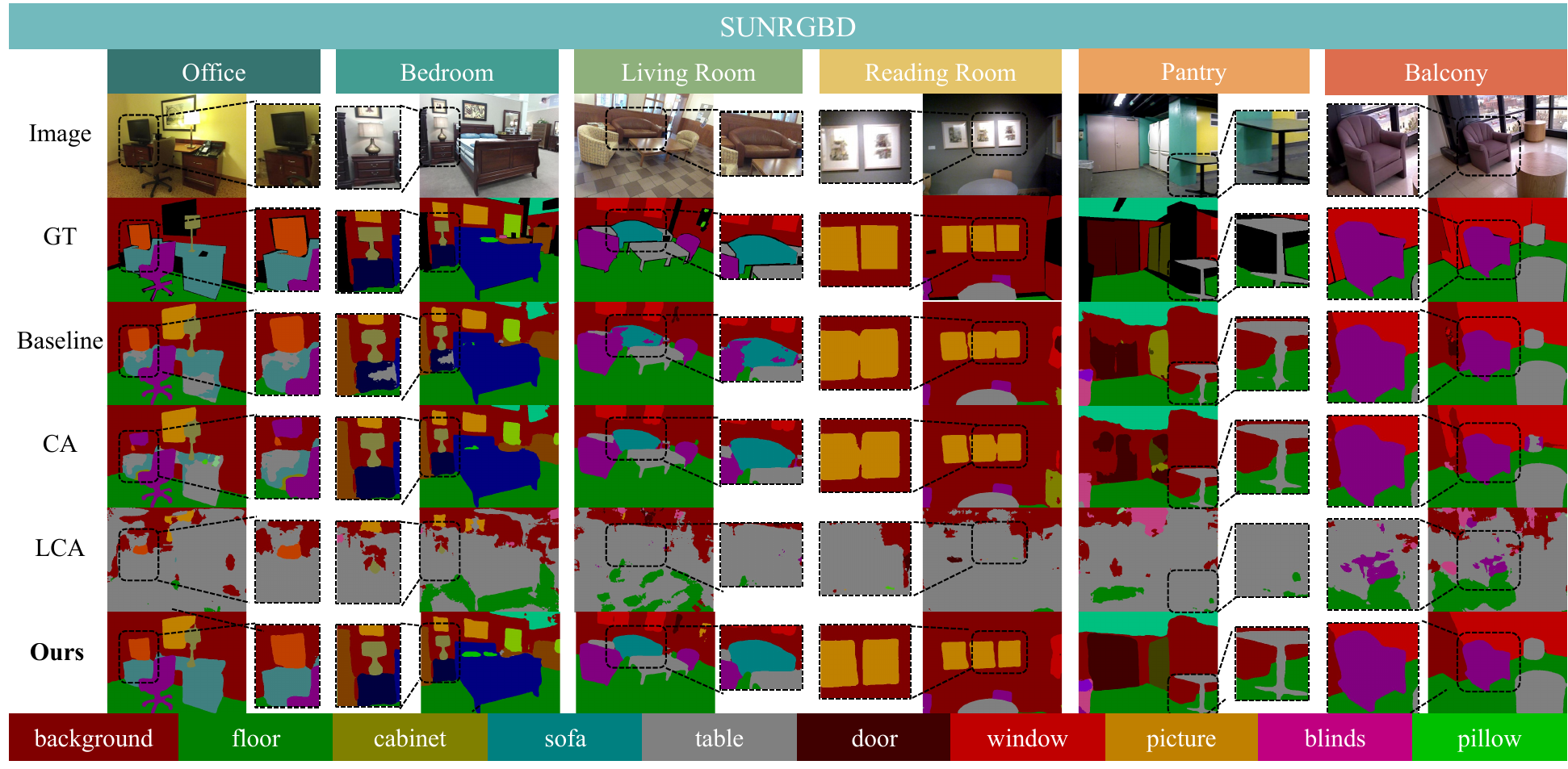}
\caption{Quantitative comparison of our DiffPixelFormer with the baseline and various cross-attention methods on SUNRGB-D.}
\label{results_sunrgb}
\end{figure*}

To further assess the effectiveness of the proposed cross-modal interaction mechanism, we perform a visual comparison on the SUN RGB-D~\cite{song2015sun} and NYUDv2~\cite{silberman2012indoor} datasets against several representative attention mechanisms, including the Baseline, Cross-Attention~\cite{vaswani2017attention}, and Local Cross-Attention~\cite{dong2022cswin}, as shown in Figs.~\ref{results_nyudv2} and~\ref{results_sunrgb}. In the six representative indoor scenes from SUN RGB-D and NYUDv2, the Baseline model exhibits blurred boundaries and frequent mis-segmentation in complex environments. Cross-Attention strengthens global dependencies between modalities but remains susceptible to background interference, while Local Cross-Attention refines local structural details yet lacks comprehensive global semantic modeling. In contrast, the proposed DiffPixelFormer attains a well-balanced trade-off between global coherence and local precision, delivering more accurate and semantically consistent segmentation, particularly in key regions such as tables, doors, and walls. These results clearly demonstrate the superiority of the proposed mechanism in capturing cross-modal complementary information and preserving semantic structural integrity.

\subsection{Ablation Studies} \label{IV-E}

\subsubsection{Effect of Overall Network Architecture}
To systematically assess the contributions of the key components to intra- and inter-modal interactions, ablation studies are conducted on the SUN RGB-D and NYUDv2 datasets (see Table~\ref{table_2}). The results show that using only intra-modal self-attention yields 51.02\% and 52.83\% mIoU on SUN RGB-D and NYUDv2, respectively. With the introduction of the cross-modal pixel-level attention module, the mIoU increases to 51.49\% and 53.57\%, indicating that explicit cross-modal interaction facilitates stronger feature association. Further adding the similarity discriminator brings the performance to 51.98\% and 54.36\%, suggesting that aligning features across modalities is beneficial for fusion. 
Incorporating the difference discriminator achieves 52.62\% and 56.09\%, highlighting the necessity of modeling modality-specific differences. Finally, with the learnable factor mechanism, DiffPixelFormer attains \textbf{52.84\%} and \textbf{56.28\%} mIoU, corresponding to \textbf{1.82\%} and \textbf{3.45\%} gains over the baseline with consistent improvements in Pixel Accuracy and mAcc. These results verify that the proposed modules provide complementary advantages and jointly enhance multi-modal semantic segmentation.

\begin{table*}[t]
\centering
\caption{Effect of Overall Network Architecture on SUN RGB-D and NYUDv2, where SA and PACA denote self-attention and pixel-aware cross-attention, respectively. The Similarity Discriminator and Difference Discriminator are integral components of DSIM. \textcolor{red}{Red} indicates improvement.}
\setlength{\tabcolsep}{3pt}
\label{table_2}
\resizebox{\textwidth}{!}{
\begin{tabular}{ccccc lcc lcc}
\toprule
\multirow{2}{*}{\makecell[c]{\textbf{SA}}} &
\multirow{2}{*}{\makecell[c]{\textbf{PACA}}} &
\multirow{2}{*}{\makecell[c]{\textbf{Similarity Discriminator}}} &
\multirow{2}{*}{\makecell[c]{\textbf{Difference Discriminator}}} &
\multirow{2}{*}{\makecell[c]{\textbf{Learning Factor}}} &
\multicolumn{3}{c}{\textbf{SUN RGB-D}} &
\multicolumn{3}{c}{\textbf{NYUDv2}} \\
\cmidrule(lr){6-8} \cmidrule(lr){9-11}
 &  &  &  &  &
\textbf{mIoU} & \textbf{Pixel Acc} & \textbf{mAcc} &
\textbf{mIoU} & \textbf{Pixel Acc} & \textbf{mAcc} \\
\midrule
\checkmark &  &  &  &  & \cellcolor{gray!15}51.02 & 82.73 & 61.92 & \cellcolor{gray!15}52.83 & 77.63 & 63.92 \\
\checkmark & \checkmark &  &  &  & \cellcolor{gray!15}51.49\textit{\fontsize{6}{0}\selectfont\textcolor{red}{\textbf{+0.47}}} & 82.86 & 63.48 & \cellcolor{gray!15}53.57\textit{\fontsize{6}{0}\selectfont\textcolor{red}{\textbf{+0.74}}} & 78.18 & 65.31 \\
\checkmark & \checkmark & \checkmark &  &  & \cellcolor{gray!15}51.98\textit{\fontsize{6}{0}\selectfont\textcolor{red}{\textbf{+0.49}}} & 83.02 & 63.94 & \cellcolor{gray!15}54.36\textit{\fontsize{6}{0}\selectfont\textcolor{red}{\textbf{+0.79}}} & 78.59 & 66.63 \\
\checkmark & \checkmark & \checkmark & \checkmark &  & \cellcolor{gray!15}52.62\textit{\fontsize{6}{0}\selectfont\textcolor{red}{\textbf{+0.64}}} & 83.31 & 64.56 & \cellcolor{gray!15}56.09\textit{\fontsize{6}{0}\selectfont\textcolor{red}{\textbf{+1.73}}} & 79.46 & 68.79 \\
\rowcolor{gray!20}
\checkmark & \checkmark & \checkmark & \checkmark & \checkmark & \textbf{52.84}\textit{\fontsize{6}{0}\selectfont\textcolor{red}{\textbf{+0.22}}} & \textbf{83.37} & \textbf{64.83} & \textbf{56.28}\textit{\fontsize{6}{0}\selectfont\textcolor{red}{\textbf{+0.19}}} & \textbf{79.79} & \textbf{68.98} \\
\bottomrule
\end{tabular}%
}
\end{table*}

\begin{table*}[t]
\large 
\centering
\caption{Effect of Different Attention Mechanisms on SUN RGB-D and NYUDv2 datasets, where SW Cross-Attention denotes the Shifted Window Cross-Attention mechanism.}
\setlength{\tabcolsep}{2pt}
\label{table_3}
\renewcommand{\arraystretch}{1.2}
\resizebox{\textwidth}{!}{
\begin{tabular}{lcccccccccccc}
\toprule
\multirow{2}{*}{\textbf{Methods}} & \multirow{2}{*}{\textbf{Backbone}} & \multirow{2}{*}{\textbf{Param(M)}} & \multicolumn{5}{c}{\textbf{SUN RGB-D}} & \multicolumn{5}{c}{\textbf{NYUDv2}} \\
\cmidrule(lr){4-8} \cmidrule(lr){9-13}
& & & \textbf{Input size} & \textbf{FLOPs (G)} & \textbf{mIoU} & \textbf{Pixel Acc} & \textbf{mAcc} & \textbf{Input size} & \textbf{FLOPs (G)} & \textbf{mIoU} & \textbf{Pixel Acc} & \textbf{mAcc} \\
\midrule
Baseline & MiT-B3 & 44.65 & 530$\times$730 & 135.58 & \cellcolor{gray!15}51.02 & 82.73 & 61.92 & 480$\times$640 & 100.75 & \cellcolor{gray!15}52.83 & 77.63 & 63.92 \\
Cross-Attention~\cite{vaswani2017attention} & MiT-B3 & 527.98 & 530$\times$730 & 749.01 & \cellcolor{gray!15}49.69 & 82.30 & 61.27 & 480$\times$640 & 482.78 & \cellcolor{gray!15}50.45 & 76.31 & 62.60 \\
SW Cross-Attention~\cite{liu2021swin} & MiT-B3 & 88.09 & 530$\times$730 & 213.19 & \cellcolor{gray!15}49.13 & 81.95 & 61.06 & 480$\times$640 & 165.37 & \cellcolor{gray!15}48.51 & 74.97 & 61.98 \\
Local Cross-Attention~\cite{dong2022cswin} & MiT-B3 & 66.38 & 530$\times$730 & 174.38 & \cellcolor{gray!15}49.81 & 82.39 & 60.56 & 480$\times$640 & 133.06 & \cellcolor{gray!15}48.80 & 75.23 & 62.42 \\
Pixel-wise Cross-Attention~\cite{jia2024geminifusion} & MiT-B3 & 66.37 & 530$\times$730 & 166.79 & \cellcolor{gray!15}51.49 & 82.86 & 63.48 & 480$\times$640 & 124.79 & \cellcolor{gray!15}53.57 & 78.18 & 65.31 \\
\rowcolor{gray!20}
\textbf{DSIM (ours)} & MiT-B3 & 85.40 & 530$\times$730 & 205.73 & \textbf{52.84} & \textbf{83.37} & \textbf{64.83} & 480$\times$640 & 154.78 & \textbf{56.28} & \textbf{79.79} & \textbf{68.98} \\
\bottomrule
\end{tabular}%
}
\end{table*}

\begin{table*}[t]
\centering
\caption{Effect of different backbones on SUN RGB-D and NYUDv2 datasets.}
\setlength{\tabcolsep}{3.5pt}
\label{table_5}
\resizebox{\textwidth}{!}{%
\begin{tabular}{lccccccccccc}
\toprule
\multirow{2}{*}{\textbf{Backbone}} & \multirow{2}{*}{\textbf{Parameters(M)}} & \multicolumn{5}{c}{\textbf{SUN RGB-D}} & \multicolumn{5}{c}{\textbf{NYUDv2}} \\
\cmidrule(lr){3-7} \cmidrule(lr){8-12}
 &  & \textbf{Input size} & \textbf{FLOPs(G)} & \textbf{mIoU} & \textbf{Pixel Acc} & \textbf{mAcc} & \textbf{Input size} & \textbf{FLOPs(G)} & \textbf{mIoU} & \textbf{Pixel Acc} & \textbf{mAcc} \\
\midrule
MiT-B1~\cite{xie2021segformer} & 25.28 & 530$\times$730 & 131.67 & \cellcolor{gray!15}47.03 & 80.93 & 59.20 & 480$\times$640 & 100.08 & \cellcolor{gray!15}32.60 & 66.10 & 44.01 \\
MiT-B3~\cite{xie2021segformer} & 85.41 & 530$\times$730 & 205.73 & \cellcolor{gray!15}52.84 & 83.37 & 64.83 & 480$\times$640 & 154.78 & \cellcolor{gray!15}56.28 & 79.79 & 68.98 \\
MiT-B5~\cite{xie2021segformer} & 157.24 & 530$\times$730 & 758.40 & \cellcolor{gray!15}53.69 & 83.47 & \textbf{66.11} & 480$\times$640 & 569.20 & \cellcolor{gray!15}58.71 & 80.30 & 69.93 \\
Swin Tiny~\cite{liu2021swin} & 51.94 & 530$\times$730 & 280.35 & \cellcolor{gray!15}49.69 & 81.84 & 62.81 & 480$\times$640 & 219.32 & \cellcolor{gray!15}46.02 & 73.51 & 59.33 \\
\rowcolor{gray!20}
Swin Large~\cite{liu2021swin} & 369.22 & 530$\times$730 & 1005.65 & \textbf{54.28} & \textbf{84.14} & 65.91 & 480$\times$640 & 815.10 & \textbf{59.95} & \textbf{81.70} & \textbf{72.74} \\
\bottomrule
\end{tabular}%
}
\end{table*}

\begin{table}[t]
\centering
\caption{Effect of Different Relation Discriminators on NYUDv2 dataset.}
\setlength{\tabcolsep}{3pt}
\label{table_4}
\resizebox{\columnwidth}{!}{%
\begin{tabular}{lcccc}
\toprule
\textbf{Relation Discriminator} & \textbf{Backbone} & \textbf{mIoU} & \textbf{Pixel Acc} & \textbf{mAcc} \\
\midrule
ECANet~\cite{wang2020eca} & MiT-b3 & \cellcolor{gray!15}53.11 & 77.92 & 66.29 \\
SENet~\cite{hu2018squeeze} & MiT-b3 & \cellcolor{gray!15}53.98 & 78.18 & 67.46 \\
2*MLP+gMLP+Softmax \cite{liu2021pay} & MiT-b3 & \cellcolor{gray!15}53.30 & 77.91 & 66.80 \\
2*MLP+Sigmoid & MiT-b3 & \cellcolor{gray!15}54.58 & 78.30 & 67.42 \\
\rowcolor{gray!20}
2*MLP+Softmax & MiT-b3 & \textbf{56.28} & \textbf{79.79} & \textbf{68.98} \\
\bottomrule
\end{tabular}%
}
\end{table}

\subsubsection{Effect of Different Attention Mechanisms}
To validate the proposed DSIM, we compare it with standard cross-attention~\cite{vaswani2017attention}, shifted window cross-attention~\cite{liu2021swin}, local cross-attention~\cite{dong2022cswin}, and pixel-wise cross-attention~\cite{jia2024geminifusion} (see Table~\ref{table_3}). Standard cross-attention, despite 527.98M parameters, suffers from diluted pixel-level details and redundancy, achieving only 49.69\% and 50.45\% mIoU on SUN RGB-D and NYUDv2. Shifted window and local variants reduce computation but their limited receptive fields constrain precise pixel alignment. Pixel-wise cross-attention enhances fine-grained inter-modal fusion, improving mIoU to 51.49\% and 53.57\%. Building on this, DSIM explicitly models differential features to capture cross-modal discrepancies, reaching 52.84\% and 56.28\% mIoU with further gains in Pixel ACC and mAcc. Compared to conventional cross-attention mechanisms, DSIM reduces parameter count and FLOPs by 83.83\% and 72.53\%, respectively. Overall, by effectively modeling cross-modal complementarity and distinctiveness, DiffPixelFormer achieves an optimal balance between accuracy and efficiency, while maintaining real-time inference at 41.66 FPS.

\subsubsection{Effect of Different Backbone Architectures}
To evaluate the influence of backbone networks on multimodal segmentation, we conducted extensive experiments on SUN RGB-D and NYUDv2, as shown in Table~\ref{table_5}. Results show that performance scales with model capacity and computational cost. The lightweight MIT-B1 with 25.28M parameters achieves only 47.03\% and 32.60\% mIoU, revealing limited ability to capture complex multimodal semantics. Increasing capacity, MIT-B3 with 85.41M parameters boosts mIoU to 52.84\% and 56.28\%, while MIT-B5 further improves to 53.69\% and 58.71\% at higher cost. 
Swin Large, with 369.22M parameters and hierarchical window-based self-attention, achieves 54.28\% and 59.95\% mIoU and 84.14\% and 81.70\% pixel accuracy, underscoring the importance of multi-scale and long-range modeling in RGB-D segmentation.
However, these gains incur substantial computational demands, underscoring trade-offs in resource-limited scenarios. Overall, our method demonstrates strong architectural compatibility, seamlessly integrating with diverse backbones to balance performance and efficiency.

\subsubsection{Effect of Different Relation Discriminators}
To assess the effectiveness of the proposed relation discriminator and identify the optimal design for cross-modal feature integration, we compare several representative approaches, including ECANet~\cite{wang2020eca}, SENet~\cite{hu2018squeeze}, and various MLP variants, as shown in Table~\ref{table_4}. Channel attention methods such as ECANet and SENet achieve 53.11\% and 53.98\% mIoU with the MiT-b3 backbone, effectively modeling channel dependencies but lacking explicit pixel-level cross-modal discrimination. The 2MLP+gMLP+Softmax variant attains 53.30\% mIoU, yet global token mixing introduces redundancy that weakens fine-grained distinctions. Replacing gMLP with a Sigmoid-based discriminator improves results to 54.58\% mIoU, confirming that independent pixel-wise discrimination better preserves local differences.
Our 2MLP+Softmax design delivers superior performance, highlighting the critical role of explicit pixel-level relation modeling and the effectiveness of the Softmax-based discriminator for fine-grained multimodal segmentation.

\section{Conclusion}
In this paper, we propose DiffPixelFormer, a novel RGB-D fusion framework whose core Intra-Inter Modal Interaction Block (IIMIB) jointly enhances intra-modal representations and models inter-modal interactions within a unified architecture. To explicitly distinguish modality-specific and shared information, IIMIB incorporates the Differential–Shared Inter-Modal (DSIM) module, which leverages pixel-level attention guided by differential and similarity cues to achieve fine-grained cross-modal alignment.
In addition, a dynamic adaptive fusion strategy is introduced to flexibly adjust modality weights according to scene characteristics, thereby fully leveraging the complementary strengths of RGB and depth data. 
DSIM reduces parameter count and FLOPs by 83.83\% and 72.53\% compared to conventional cross-attention mechanisms. Extensive experiments on NYUDv2 and SUN RGB-D demonstrate that DiffPixelFormer outperforms state-of-the-art methods while maintaining real-time inference at 41.66 FPS.
In future work, we will extend the proposed differential pixel-aware mechanism to broader multimodal perception tasks and explore its generalization and application potential under modality-missing scenarios.

\bibliographystyle{IEEEtran}
\bibliography{mybib}

\vfill

\end{document}